\documentclass[runningheads]{llncs}

\usepackage{graphicx}
\usepackage{booktabs}
\usepackage{multirow}
\usepackage[final]{eccv}
\usepackage[pagebackref,breaklinks,colorlinks,bookmarks=false]{hyperref}

\definecolor{findingframe}{RGB}{38,148,132}
\definecolor{findingback}{RGB}{233,247,244}
\newsavebox{\findingbox}
\newenvironment{finding}
  {\par\smallskip\noindent\begin{lrbox}{\findingbox}%
   \begin{minipage}{\dimexpr\linewidth-2\fboxsep-2\fboxrule\relax}\small}
  {\end{minipage}\end{lrbox}%
   \setlength{\fboxrule}{0.9pt}%
   \noindent\fcolorbox{findingframe}{findingback}{\usebox{\findingbox}}\par\smallskip}

\begin{document}

\title{Attributes Should Come from Images, Not Class Names: Distribution-Conditioned Attribute Selection for Vision-Language Models}

\titlerunning{Attributes Should Come from Images, Not Class Names}


\author{Gautam Rajendrakumar Gare \and Jia Shi \and Zhiqiu Lin \and Deepak Pathak \and\\ John Galeotti \and Deva Ramanan}

\authorrunning{Gare et. al.}

\institute{Carnegie Mellon University, USA}

\maketitle

\begin{abstract}
A popular route to interpretable zero-shot classification asks a large language model (LLM) to describe each class name and prompts CLIP with the resulting descriptors.
We show that these descriptors carry little visual evidence of their own: removing the class name from the prompt collapses ImageNet accuracy from 59.5\% to 15.5\%.
The diagnosis is that the descriptors are conditioned on the label rather than on the images, so they describe the concept in general and mislead exactly when the data shifts; an LLM insists that strawberries are red, but every strawberry in ImageNet-Sketch is a colorless line drawing.
We therefore select attributes from the target image collection instead: we score a large attribute pool against the images in CLIP's joint embedding space and keep the top-scoring attributes per class.
Selected this way, class-name-free attribute prompts reach 23.8\% on ImageNet (against 15.5\% for LLM descriptors), the gain holds on four shifted ImageNet variants, and reselecting from the LLM's own pool isolates the selection mechanism as the cause.
With one image per class, the selected attributes outperform the prompt-tuning method CoOp by 3 points while fitting in under a minute instead of 14 hours, with no learned soft prompt to obscure the decision.
Because the attribute set is chosen by the data, it doubles as a readable summary of a dataset, which we use to describe distribution shift in words.
Our code and results are available on our project page: \url{https://ggare-cmu.github.io/AttributeSelect/}.
\end{abstract}

\section{Introduction}

\begin{figure}[t]
\begin{center}
   \includegraphics[width=\linewidth]{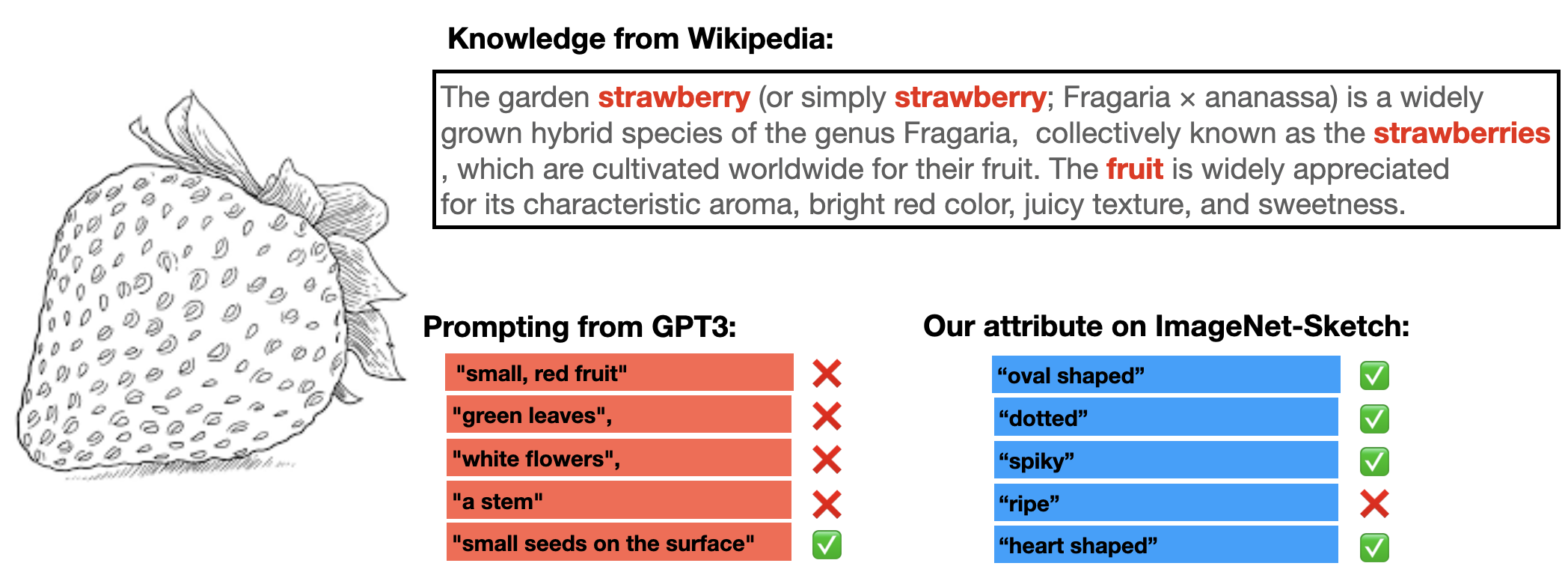}
\end{center}
    \vspace{-2em}
   \caption{
   \textbf{Class-name-conditioned descriptors fail under distribution shift; image-conditioned attributes remain reliable.}
    For the class \emph{strawberry} on ImageNet-Sketch, an LLM prompted only with the class name~\cite{menon2023visual} or an external knowledge base~\cite{Shen2022K-LITE:Knowledge} produces canonical descriptors (e.g., \emph{red}, \emph{green leaves}) because neither ever sees an image. In contrast, our method scores a large pool of candidate attributes against the sketch images themselves, selecting semantic attributes that remain valid in the shifted distribution (e.g., \emph{heart-shaped}, \emph{dotted}) thus aligning them with distribution-specific cues such as \emph{colorless} and \emph{hand-drawn}, rather than relying on source-domain appearance priors.
Section~\ref{sec:confound} quantifies the failure; Section~\ref{sec:distshift} turns the same mechanism into a description of the shift.
   %
   }
\label{fig:strawberry}
\end{figure}

Prompting a vision-language model (VLM) such as CLIP~\cite{radford2021clip} with class names turns it into a zero-shot classifier.
A popular refinement asks an LLM to describe each class (``a strawberry is red, has seeds, ...'') and averages the prompts built from these descriptors~\cite{menon2023visual,Pratt2022WhatClassification}, improving accuracy and offering an interpretation: the image was classified by its attributes.
The descriptors, however, are generated from the class name alone; no image is ever consulted.

This paper starts from a direct test of what those descriptors measure.
When we score the descriptors of Menon and Vondrick~\cite{menon2023visual} without the class name in the prompt, ImageNet accuracy collapses from 59.5\% to 15.5\% (Table~\ref{tab:main}).
The descriptors themselves supply little visual evidence; the class name does the work.
The same conditioning failure is visible qualitatively under distribution shift: the LLM describes strawberries as red and ripe, but every strawberry in ImageNet-Sketch is a colorless line drawing (Figure~\ref{fig:strawberry}), so the descriptors point the classifier the wrong way precisely when the class name needs help.

Our diagnosis is that descriptors are conditioned on the wrong variable.
A class name is a pointer to a concept, and an LLM can only describe the concept in general; the images at hand may look nothing like the general case.
The fix follows from the diagnosis: condition attribute selection on the images.
We score a large pool of candidate attributes against the target image collection in CLIP's joint embedding space and keep, per class, the attributes the images themselves rank highest.
Like Menon and Vondrick~\cite{menon2023visual}, we classify with attribute prompts in the joint space of a frozen VLM; unlike them, we select attributes from the images rather than generating them from the class name, which matters because the attribute set then tracks the distribution it describes (Table~\ref{tab:main}, bottom block).

Conditioning on images pays off beyond the diagnostic.
Attribute selection needs so few labeled images that it becomes a strong few-shot learner: with one or two images per class, the selected attributes outperform gradient-based prompt tuning while remaining human-readable and fitting in seconds.
And because the selected attributes describe the data rather than the label, the mean attribute profile of a dataset is itself an interpretable object: subtracting the profiles of two datasets describes their distribution shift in words.

Our contributions are threefold.
First, we establish the class-name confound: LLM-generated descriptors lose most of their accuracy once the class name is removed from the prompt, so their apparent gains cannot be read as attribute-level evidence (Section~\ref{sec:confound}).
Second, we propose distribution-conditioned attribute selection, a training-light procedure that scores a fixed attribute pool against the target images with a frozen CLIP and keeps the top-ranked attributes per class; a matched-pool comparison attributes the resulting gains to the selection mechanism rather than the vocabulary (Sections~\ref{sec:method} and~\ref{sec:results}).
Finally, we show that the selected attributes act as an interpretable feature space: they support extreme few-shot classification, describe distribution shift in words, extend CLIP to vocabulary its text encoder cannot parse, and localize objects without class names (Sections~\ref{sec:fewshot} and~\ref{sec:applications}).

\begin{figure}[t]
\begin{center}
 \includegraphics[width=\linewidth]{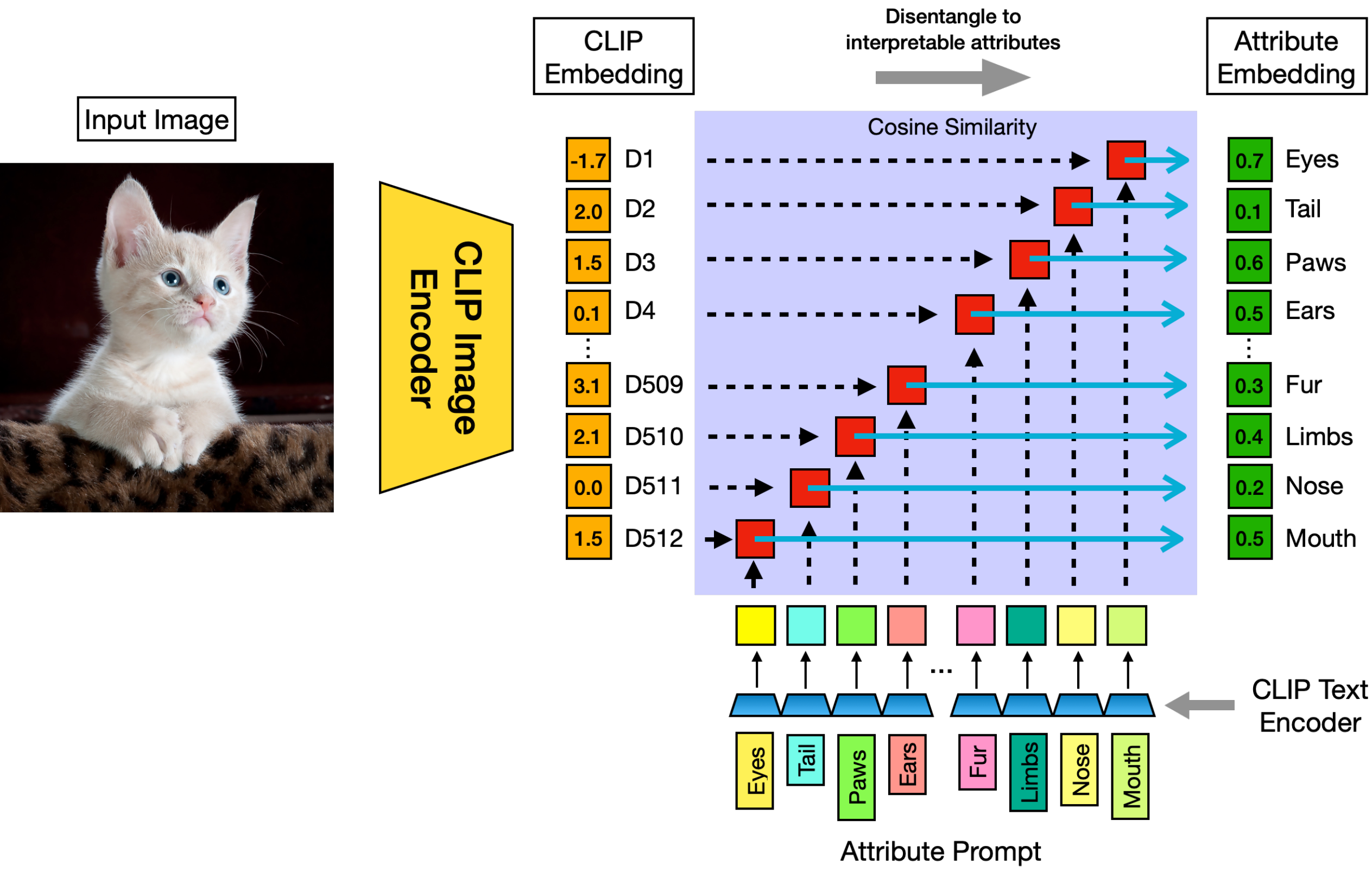}
\end{center}
    \vspace{-2em}
   \caption{\textbf{Distribution-conditioned attribute selection.}
   A frozen CLIP encodes the target images and a large pool of attribute texts; the cosine similarities between them form an attribute feature vector per image.
   Class-attribute weights are the per-class mean of these features (or the weights of a linear probe trained on them), and the top-$k$ attributes per class define the class's prompt set.
   Both encoders stay frozen throughout; the only fitted object is the linear pairing.}
\label{fig:overview}
\end{figure}

\section{Related Work}

{\bf Descriptor prompting with LLMs.}
Menon and Vondrick~\cite{menon2023visual} prompt GPT-3 for per-class descriptors and average the resulting prompts; CuPL~\cite{Pratt2022WhatClassification} generates full prompt sentences the same way; K-LITE~\cite{Shen2022K-LITE:Knowledge} draws external knowledge from WordNet and Wiktionary.
All three condition the added text on the class name alone.
Roth et al.~\cite{Roth2023WafflingConcepts} showed that replacing such descriptors with random words matches their accuracy once the class name is present, questioning what the descriptors contribute.
Our class-name-free evaluation is complementary: we measure the descriptors alone and show they collapse, then repair them by conditioning the selection on images rather than on the label.

{\bf Concept bottlenecks and interpretable embeddings.}
Concept bottleneck models route the prediction through human-named concept scores~\cite{koh2020concept}; LaBo~\cite{yang2023labo} and Label-Free CBM~\cite{Oikarinen2023Label-FreeModels} remove the concept-annotation requirement by generating candidate concepts with an LLM and scoring them with CLIP, and SpLiCE~\cite{Bhalla2024InterpretingSpLiCE} decomposes CLIP embeddings into sparse combinations of concept vectors.
Architecturally our classifier is a label-free concept bottleneck: attribute scores feed a linear layer.
The difference is where the concepts come from; the LLM-generated pools of~\cite{yang2023labo,Oikarinen2023Label-FreeModels} inherit the class-name conditioning we diagnose in Section~\ref{sec:confound}, whereas we select concepts with the target images and show that this choice, holding the pool fixed, is what moves accuracy.

{\bf Attribute datasets and prompt tuning.}
Attributes have a long history in zero-shot recognition~\cite{lampert2009learning,akata2013label,farhadi2009describing,Jayaraman2014ZeroAttributes}; we use the modern large-scale pools VAW~\cite{Pham2021LearningWild} and LSA~\cite{Pham2022ImprovingTransformers} as our candidate vocabulary.
On the adaptation side, CoOp~\cite{zhou2022coop} tunes continuous prompt vectors by gradient descent and WiSE-FT~\cite{wortsman2022robust} ensembles fine-tuned and zero-shot weights; both are strong few-shot baselines but neither yields a human-readable decision.
We compare against both in Section~\ref{sec:fewshot}.

\section{The Class-Name Confound}
\label{sec:confound}

{\bf Descriptor accuracy rides the class name.}
The standard protocol of~\cite{menon2023visual} prompts CLIP with ``\{class name\}, which \{has/is\} \{descriptor\}'' and averages over each class's descriptors.
On ImageNet with CLIP-RN50 this reaches 59.5\%, ahead of the 55.3\% class-name-only baseline (Table~\ref{tab:main}, top block).
The interpretability claim implicit in this protocol is that the descriptors measure attribute evidence in the image.
If that were true, the descriptors should retain much of their accuracy when queried alone.
They do not: with the template reduced to ``\{descriptor\}'', the same descriptor set scores 15.5\% (Table~\ref{tab:main}, bottom block), a 44-point collapse, and the pattern repeats on every shifted ImageNet variant we test.
Roth et al.~\cite{Roth2023WafflingConcepts} reached a consistent conclusion from the opposite direction: with the class name kept, random words do as well as LLM descriptors.

{\bf Why the conditioning is wrong.}
The descriptors are a function of the class name only, so they can at best describe the concept's typical appearance.
Whenever the target distribution departs from typical, label-conditioned descriptors are not merely uninformative but wrong: on ImageNet-Sketch, ``red'' and ``ripe'' actively vote against every strawberry in the dataset (Figure~\ref{fig:strawberry}).
Two explanations for the collapse come to mind that do not indict the conditioning: CLIP might simply be unable to score attribute presence without a class anchor, or the GPT-generated pool might be too small and generic to describe images.
Section~\ref{sec:attronly} tests and rejects both by reselecting from the very same pool.

\begin{finding}\textbf{Finding.} LLM-generated descriptors ride the class name: removing it from the prompt collapses ImageNet accuracy from 59.5\% to 15.5\%. Their gains cannot be read as attribute-level evidence about the image.
\end{finding}

\section{Method: Selecting Attributes with Images}
\label{sec:method}

The diagnosis prescribes the fix: attributes must be conditioned on the images they are meant to describe.
We keep the pool of candidate attributes fixed and let the target image collection rank them, using nothing but a frozen CLIP.
Figure~\ref{fig:overview} summarizes the pipeline.

{\bf Attribute scoring.}
Given an image $I$, the CLIP image encoder yields an embedding $v = f_{\text{img}}(I) \in \mathbb{R}^{n}$.
Given a pool of $m$ attribute strings $\{a_1, \dots, a_m\}$, the text encoder yields embeddings $t_j = f_{\text{txt}}(a_j)$.
The attribute feature of the image is the vector of scaled cosine similarities
\begin{equation}
 s_j = \tau \cdot \cos(v, t_j), \qquad j = 1, \dots, m,
\end{equation}
with temperature $\tau = 100$ following~\cite{radford2021clip}.
We encode each attribute string bare, with no template and no class name.
This choice is deliberate: appending the class name would correlate all of a class's attribute scores through the shared name, and it is exactly the confound of Section~\ref{sec:confound} that we need to avoid.

{\bf Class-attribute pairing.}
Given images with class labels, we estimate a score $w_{c,j}$ for how well attribute $j$ describes class $c$, in one of two ways.
The simplest sets $w_{c,j}$ to the mean of $s_j$ over the images of class $c$; this needs no training at all and drives the distribution-level applications of Section~\ref{sec:applications}.
The classification experiments instead train a linear probe on the attribute features and use its per-class weights (plus bias) as $w_{c,j}$, which sharpens the estimate when labels are available.
Either way, class $c$ keeps the $k$ attributes with the largest scores; this ranked word list $A_c$ is the class's prompt set.

{\bf Classifying with selected attributes.}
A test image with embedding $v$ is assigned to the class whose selected attributes it matches best:
\begin{equation}
\hat{c} = \arg\max_{c} \; \frac{1}{|A_c|} \sum_{j \in A_c} \omega_{c,j} \cdot \cos(v, t_j),
\label{eq:score}
\end{equation}
where $\omega_{c,j} = 1$ by default, so every selected attribute votes equally; the \emph{weighted} variant of Section~\ref{sec:fullpool} sets $\omega_{c,j} = w_{c,j}$ instead, so the attributes most characteristic of a class carry more of its vote.
Depending on the protocol under evaluation, each attribute is encoded bare or appended to the class name using the template of ~\cite{menon2023visual}.

{\bf The attribute pool.}
We pool attributes from VAW~\cite{Pham2021LearningWild}, LSA~\cite{Pham2022ImprovingTransformers}, and the GPT-3 descriptors of~\cite{menon2023visual}, and preprocess the strings to remove explicit class names, duplicates, and punctuation (details in the supplementary material).
Everything runs with both CLIP encoders frozen; no image-text pair is ever used for training.
This puts the method at the opposite end of the cost spectrum from attribute-supervised VLMs such as STAIR~\cite{chen2023stair}, which trains on over a billion pairs to learn an attribute vocabulary.

\section{Results}
\label{sec:results}

\subsection{Setup}
All experiments use CLIP with the ResNet-50 backbone~\cite{radford2021clip,He2016DeepRecognition} unless stated otherwise.
We evaluate top-1 accuracy on ImageNet and four shifted variants: ImageNetV2~\cite{Recht2019DoImageNet}, ImageNet-Sketch~\cite{Wang2019LearningPower}, ImageNet-A~\cite{hendrycks2021natural}, and ImageNet-R~\cite{hendrycks2021many}.
Attributes are always selected on ImageNet training images only; the shifted variants are never seen during selection.
Attribute-pool preprocessing and full training details are in the supplementary material.

\begin{table}[t]
\centering
\renewcommand{\arraystretch}{1.1}
\resizebox{\linewidth}{!}{
\begin{tabular}{l|ccccc}
\toprule
Prompt & ImageNet & -V2 & -Sketch & -A & -R \\
\midrule
\multicolumn{6}{l}{\emph{Class name in the prompt}} \\
\midrule
Class name only (zero-shot CLIP) & 55.31 & 49.39 & 31.58 & 20.92 & 58.10 \\
+ LLM descriptors~\cite{menon2023visual} & 59.47 & 52.84 & 33.75 & 23.51 & 57.32 \\
+ Ours (top 5) & 59.53 & 53.33 & 33.12 & 22.79 & 56.30 \\
+ Ours (top 10) & 60.18 & 53.40 & 33.66 & 23.43 & 57.49 \\
+ Ours (top 100) & {\bf 60.80} & {\bf 53.95} & {\bf 34.27} & {\bf 24.00} & {\bf 59.04} \\
\midrule
\multicolumn{6}{l}{\emph{Attribute only (no class name)}} \\
\midrule
LLM descriptors~\cite{menon2023visual} & 15.50 & 14.00 & 9.60 & 8.57 & 17.91 \\
Ours, same pool as~\cite{menon2023visual} (top 5) & 19.40 & 17.00 & 10.57 & 9.80 & 19.49 \\
Ours, VAW+LSA pool~\cite{Pham2021LearningWild,Pham2022ImprovingTransformers} (top 5) & {\bf 23.80} & {\bf 20.80} & {\bf 14.40} & {\bf 11.83} & {\bf 28.20} \\
\bottomrule
\end{tabular}
}
\vspace{1mm}
\caption{\textbf{Attribute prompts stand on their own only when selected from images.}
Top-1 accuracy with CLIP-RN50; attributes are selected on ImageNet only and reused unchanged on the four shifted variants.
Top block: with the class name in the prompt, all descriptor methods sit within about a point of each other.
Bottom block: with the class name removed, LLM descriptors collapse (59.47 to 15.50 on ImageNet) while our selection recovers 3.9 points from the identical pool and 8.3 points from a larger pool.
{\bf Bold} marks the best result per column within each block.}
\label{tab:main}
\end{table}

\subsection{Attribute-only classification}
\label{sec:attronly}

{\bf Selection, not vocabulary, closes the gap.}
We return to Table~\ref{tab:main} (bottom block), now with our rows.
Reselecting attributes from the exact pool of~\cite{menon2023visual}, conditioned on ImageNet training images, lifts class-name-free accuracy from 15.50\% to 19.40\%.
The pool, the model, and the evaluation protocol are identical; only the selection mechanism changed, so the 3.9-point gain is attributable to conditioning on images.
This also rejects both alternative explanations from Section~\ref{sec:confound}: CLIP scores attribute presence well enough to reach 19.40\% over 1{,}000 classes with five bare words per class, and the GPT pool does contain useful attributes; label-conditioned generation simply fails to surface them.
Widening the pool to VAW+LSA raises accuracy further to 23.80\%, a 53\% relative improvement over the LLM descriptors.

{\bf The gain survives distribution shift.}
Attributes selected on ImageNet transfer unchanged to the four shifted variants and preserve the ordering everywhere, with the largest margin on ImageNet-R (28.20 vs 17.91).
Absolute numbers remain far below class-name prompting, which is expected; a handful of attributes shared across many classes cannot fully separate 1{,}000 categories, and we state this boundary in Section~\ref{sec:limitations}.
The top-5 budget is also deliberately harsh; it buys readability, not accuracy.
Uncapping it raises attribute-only ImageNet accuracy to 45.5\%, nearly three times the LLM-descriptor baseline (Section~\ref{sec:fullpool}).

\subsection{Uncapping the attribute budget}
\label{sec:fullpool}

The headline attribute-only results above restrict each class to its top-5 attributes, the harshest and most readable setting.
This section removes the cap: each class keeps its full ranked attribute list, selected from a growing number of images per class.
We compare the two scoring variants of Eq.~\eqref{eq:score}: unweighted ($\omega_{c,j} = 1$, every selected attribute votes equally) and weighted ($\omega_{c,j} = w_{c,j}$, a class's most characteristic attributes vote more).

{\bf Weights help consistently but modestly.}
Table~\ref{tab:fullpool} compares unweighted and weighted variants at 16 selection images per class.
Weighting the selected attributes by their class-attribute scores adds roughly 0.3 points on every dataset, for both the LLM descriptor pool and ours.
The pool matters far more than the weighting: the full VAW+LSA pool reaches 45.53\% on ImageNet against 18.06\% for the LLM descriptor pool under identical selection and weighting.

{\bf Attribute-only accuracy scales with selection images.}
Table~\ref{tab:shotscaling} traces the same weighted and unweighted variants from 1 to 16 selection images per class.
Accuracy rises monotonically on all five datasets (ImageNet: 23.17\% at 1 shot to 45.53\% at 16): with the budget uncapped, the selection keeps improving as more images refine the class-attribute scores.
For calibration, the LLM descriptors, which use no images, sit at 15.50\% on ImageNet; a single selection image per class already exceeds them under the full weighted pool.

 \begin{finding}
 \textbf{Finding.} Conditioning attribute selection on images, with the pool held fixed, lifts class-name-free ImageNet accuracy from 15.5\% to 19.4\%; a larger pool reaches 23.8\% with five attributes per class and 45.5\% with the full weighted pool. The selection mechanism, not the vocabulary, is what makes attributes informative.
 \end{finding}

\begin{table}[t]
\centering
\renewcommand{\arraystretch}{1.1}
\resizebox{\linewidth}{!}{
\begin{tabular}{llc|ccccc}
\toprule
Pool & Selection & Weighted & ImageNet & -V2 & -Sketch & -A & -R \\
\midrule
LLM descriptors~\cite{menon2023visual} & none (as released) & -- & 15.50 & 14.00 & 9.60 & 8.57 & 17.91 \\
\midrule
\multirow{2}{*}{LLM descriptor pool} & \multirow{2}{*}{ours, full pool} & no & 16.67 & 13.15 & 6.44 & 6.44 & 23.14 \\
& & yes & 18.06 & 14.45 & 7.08 & 6.68 & 24.16 \\
\midrule
\multirow{2}{*}{VAW+LSA} & \multirow{2}{*}{ours, full pool} & no & 45.25 & 37.25 & 19.14 & 14.04 & 36.97 \\
& & yes & {\bf 45.53} & {\bf 37.50} & {\bf 19.41} & {\bf 14.43} & {\bf 37.27} \\
\bottomrule
\end{tabular}
}
\vspace{1mm}
\caption{\textbf{With the attribute budget uncapped, image-conditioned selection triples attribute-only accuracy.}
Top-1 accuracy, attribute-only prompts (no class name), CLIP-RN50; selection uses 16 images per class.
Weighted = each attribute's vote scaled by its class-attribute score (Eq.~\eqref{eq:score} with $\omega_{c,j} = w_{c,j}$); it adds about 0.3 points everywhere, while switching from the LLM descriptor pool to VAW+LSA under identical selection adds over 27 points on ImageNet.
{\bf Bold} marks best result per column.}
\label{tab:fullpool}
\end{table}

\begin{table}[t]
\centering
\renewcommand{\arraystretch}{1.1}
\resizebox{0.72\linewidth}{!}{
\begin{tabular}{ll|ccccc}
\toprule
\multirow{2}{*}{Dataset} & \multirow{2}{*}{Weighted} & \multicolumn{5}{c}{Selection images per class} \\
\cmidrule(l){3-7}
& & 1 & 2 & 4 & 8 & 16 \\
\midrule
\multirow{2}{*}{ImageNet} & no & 22.92 & 30.36 & 36.90 & 42.06 & 45.25\\
& yes & {\bf 23.17} & {\bf 30.64} & {\bf 37.20} & {\bf 42.60} & {\bf 45.53}\\
\midrule
\multirow{2}{*}{ImageNetV2} & no & 19.20 & 26.04 & 31.32 & 35.27 & 37.25\\
& yes & {\bf 19.54} & {\bf 26.20} & {\bf 31.55} & {\bf 35.43} & {\bf 37.50} \\
\midrule
\multirow{2}{*}{ImageNet-Sketch} & no & 10.28 & 13.19 & 15.83 & 17.85 & 19.14\\
& yes & {\bf 10.41} & {\bf 13.41} & {\bf 16.08} & {\bf 18.08} & {\bf 19.41} \\
\midrule
\multirow{2}{*}{ImageNet-A} & no & 9.93 & 12.13 & 12.08 & 13.38 & 14.04 \\
& yes & {\bf 10.00} & {\bf 12.21} & {\bf 12.22} & {\bf 13.48} & {\bf 14.43}\\
\midrule
\multirow{2}{*}{ImageNet-R} & no & 25.53 & 29.63 & 32.00 & 35.55 & 36.97\\
& yes & {\bf 25.77} & {\bf 29.91} & {\bf 32.16} & {\bf 35.81} & {\bf 37.27} \\
\bottomrule
\end{tabular}
}
\vspace{1mm}
\caption{\textbf{Attribute-only accuracy grows monotonically with selection images when the attribute budget is uncapped.}
Top-1 accuracy, attribute-only prompts (no class name), full VAW+LSA pool, CLIP-RN50.
Weighted is defined as in Table~\ref{tab:fullpool}; it adds a consistent margin of roughly 0.3 points at every shot count on every dataset.
{\bf Bold} marks the better of the two variants per column within each dataset.}
\label{tab:shotscaling}
\end{table}

\subsection{Prompting with class names}
With the class name restored, our selected attributes give a consistent but small edge, with a monotone dose response in $k$: 59.53, 60.18, and 60.80 for the top 5, 10, and 100 attributes, against 59.47 for LLM descriptors (Table~\ref{tab:main}, top block), and the ordering holds on all four shifted variants.
We do not lean on this comparison: once the class name is present, much of any descriptor method's gain is prompt ensembling rather than attribute evidence~\cite{Roth2023WafflingConcepts}, which is precisely the confound this paper is about.
The class-name-free block carries our claim; the top block shows that image-conditioned selection at least matches label-conditioned generation on the standard protocol too.

\begin{table}[t]
\centering
\renewcommand{\arraystretch}{1.1}
\resizebox{\linewidth}{!}{
\begin{tabular}{l|ccccc|c}
\toprule
\multirow{2}{*}{Method} & \multicolumn{5}{c|}{Number of shots} & \multirow{2}{*}{Fit time}\\
\cmidrule(l){2-6}
& 1 & 2 & 4 & 8 & 16 & \\
\midrule
CLIP linear probe & 24.46 & 35.09 & 44.60 & 52.31 & 57.10 & $<$1 min \\
CoOp~\cite{zhou2022coop} & 57.15 & 57.81 & 59.99 & 61.56 & {\bf 62.95} & 14 hr \\
WiSE-FT~\cite{wortsman2022robust} & 58.30 & 59.08 & 60.48 & {\bf 61.85} & 62.84 & $<$1 min\\
Ours & {\bf 60.13} & {\bf 60.92} & {\bf 61.13} & 61.39 & 61.35 & $<$1 min\\
\bottomrule
\end{tabular}
}
\vspace{1mm}
\caption{\textbf{Attribute selection is a strong extreme-few-shot learner.}
ImageNet top-1 accuracy with CLIP-RN50; zero-shot CLIP scores 55.31.
We linear-probe the attribute features of the $k$-shot images to select top-$k$ attributes per class, then evaluate zero-shot with the selected attributes; baselines fine-tune or prompt-tune on the same shots.
{\bf Bold} marks the best method per column.
Our accuracy plateaus beyond 4 shots because the probe only ranks a fixed pool of words; the crossover is stated, not hidden.}
\label{tab:fewshot}
\end{table}

\subsection{Few-shot classification}
\label{sec:fewshot}

{\bf One image per class is enough to select good attributes.}
We select each class's top attributes from only $k$ labeled images and evaluate zero-shot with the selected attribute prompts.
At 1 shot this reaches 60.13\% against 57.15\% for CoOp~\cite{zhou2022coop}, with the margin persisting at 2 and 4 shots (Table~\ref{tab:fewshot}), and the fit takes under a minute against CoOp's 14 hours of prompt tuning.
Unlike a tuned soft prompt, the fitted object is a list of words per class, so the resulting classifier can be read and audited; the supplementary material discusses editing it by intervention.

{\bf Why selection wins the low-shot regime, and where it stops.}
Selection only has to rank a fixed pool of semantically meaningful candidates, a far smaller hypothesis space than a continuous prompt, so a single image carries enough signal; the attribute features' shared semantic structure (Section~\ref{sec:featstructure}) does the rest.
The same property caps the method: the probe's only job is to rank the pool, so extra images refine the ranking but add no capacity, and accuracy plateaus near 61.4\% while CoOp and WiSE-FT keep improving and pass us at 8 shots.
The method's regime is the extreme low-shot end, and Table~\ref{tab:fewshot} shows both sides of the boundary.

 \begin{finding}
 \textbf{Finding.} With one image per class, image-conditioned attribute selection beats CoOp by 3 points (60.1 vs 57.2) at $10^{-3}$ of the fitting cost, while producing a classifier that is a readable list of words; beyond 4 shots, gradient methods pass it.
\end{finding}

\subsection{Structure of the attribute features}
\label{sec:featstructure}
\vspace{-0.8em}

\begin{figure}[t]
\begin{center}
   \includegraphics[width=\linewidth]{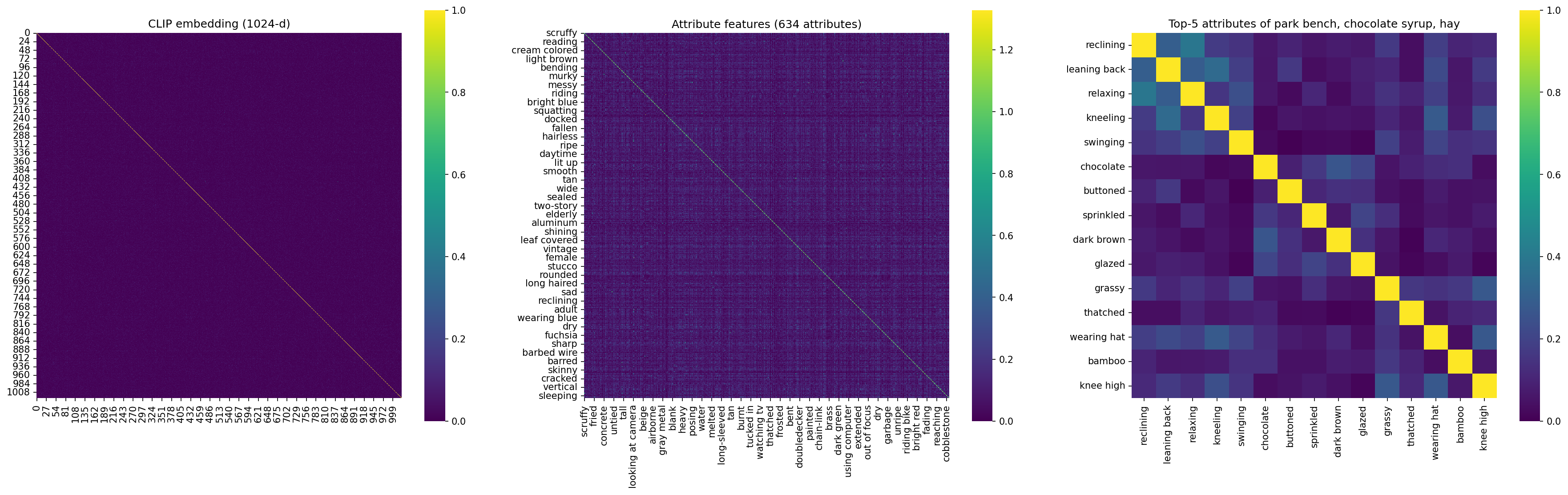}
\end{center}
    \vspace{-2em}
   \caption{\textbf{Attribute features trade statistical independence for named, semantically clustered directions.}
   Pairwise mutual information on ImageNet with CLIP-RN50; all three panels use the sequential viridis scale, dark purple at zero to bright yellow at high mutual information.
   Left: dimensions of the 1024-d CLIP embedding, near zero off the diagonal.
   Middle: our 634 VAW attribute features, visibly more correlated.
   Right: features restricted to the top-5 attributes of three classes (park bench, chocolate syrup, hay); the brighter 5$\times$5 diagonal blocks show that a class's own attributes share information, i.e.\ the correlations follow semantics.}
\label{fig:feature_disen}
\end{figure}

Figure~\ref{fig:feature_disen} compares pairwise mutual information among CLIP embedding dimensions and among our attribute features on ImageNet.
CLIP dimensions are close to statistically independent; attribute features are visibly correlated, and the top-5 attributes of a class share more information with each other than with other classes' attributes (right panel).
We read this honestly: the transformation is not statistical disentanglement, it is a change of basis into named directions whose correlations are semantic (an object that scores high on ``ripe'' also scores high on ``red'').
That named, semantically clustered structure is what the few-shot result exploits, and what every application in Section~\ref{sec:applications} depends on.

\section{Applications of Data-Conditioned Attributes}
\label{sec:applications}

The applications below share one mechanism: the attribute set is a function of an image collection, so any collection (a dataset, a class under shift, an image region) can be summarized in words.
Label-conditioned descriptors cannot do this by construction, since they never see images.

\subsection{Describing distribution shift in words}
\label{sec:distshift}
\vspace{-0.8em}

\begin{figure}[!t]
\begin{center}
   \includegraphics[width=\linewidth]{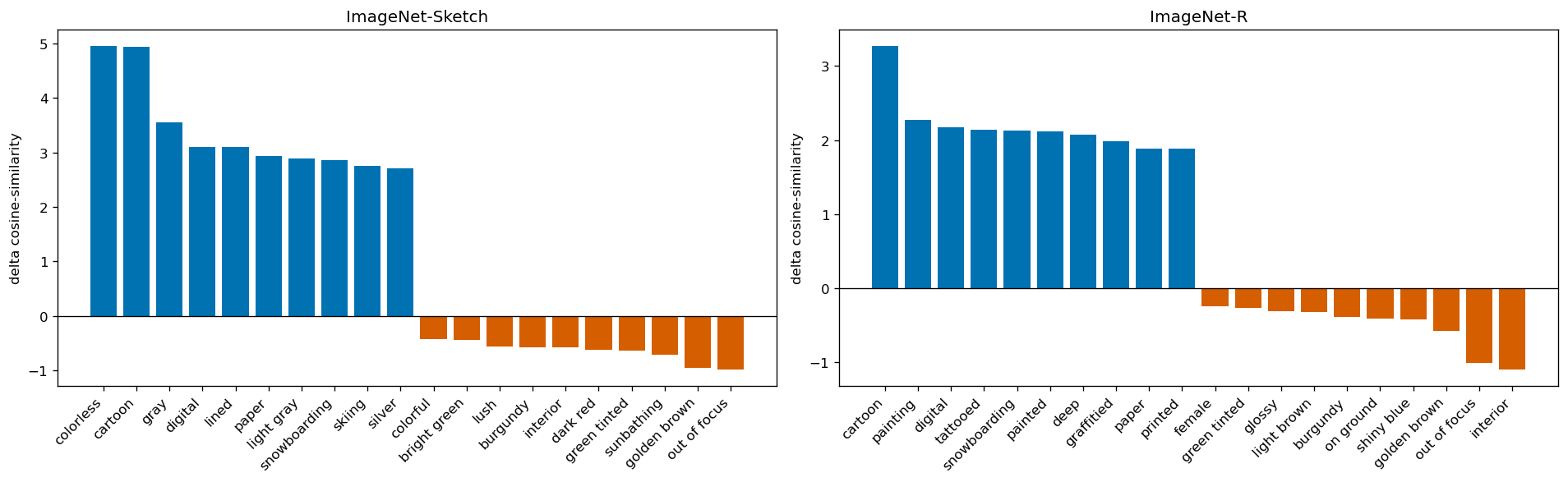}
\end{center}
    \vspace{-2em}
   \caption{\textbf{Distribution shift, described in words.}
   Top-10 rising and falling attributes (change in mean attribute score over unlabeled images, y-axis) of ImageNet-Sketch (left panel) and ImageNet-R (right panel) relative to ImageNet.
   Sketch gains ``colorless'', ``gray'', and ``cartoon'' and loses color attributes; R gains ``cartoon'', ``painting'', and ``tattooed'', matching its documented composition.
   Within each panel attributes are ordered by their change, risers first.
   Blue bars rising above zero become more present under the shift; vermillion bars falling below zero become less present.}
\label{fig:sketch_dist_shift}
\end{figure}

Subtracting the mean attribute profile of ImageNet from that of a shifted dataset ranks the attributes that became more and less present.
On ImageNet-Sketch the top risers are ``colorless'', ``gray'', ``cartoon'', and ``digital'' while color attributes fall; on ImageNet-R the risers are ``cartoon'', ``painting'', and ``tattooed'', matching that dataset's stated composition of art, cartoons, and tattoo renditions~\cite{hendrycks2021many} (Figure~\ref{fig:sketch_dist_shift}).
We use only unlabeled test images, so no class-specific information leaks.
Beyond inspection, such profiles give a quantitative, interpretable measure of domain gap, relevant to domain adaptation prompting~\cite{dunlap2023lads} and to monitoring shift in continual settings~\cite{lin2021clear}.

{\bf Domain attributes as prompt templates.}
The description is also actionable.
We take the top-5 rising attributes of each shifted dataset, computed from unlabeled images only, and build the prompt template ``A \{attribute\} photo of a \{class\}'', ensembled over the five attributes; on ImageNet-R the templates become ``A cartoon photo of ...'', ``A painting photo of ...'', and so on.
Table~\ref{tab:domaintemplate} evaluates this template against the plain ``\{class\}'' and ``A photo of a \{class\}'' templates, crossed with our class-specific attribute prompts.
The domain template improves plain zero-shot on every dataset, by up to 3.7 points on ImageNet-Sketch, and still adds up to 1.5 points on top of our strongest top-100 attribute prompts.
Domain-level and class-level attributes are therefore complementary: one describes how the images look, the other what they contain.

\begin{table}[t]
\centering
\resizebox{0.9\linewidth}{!}{
\begin{tabular}{ll|c|c|c}
\toprule
\multirow{2}{*}{Dataset} & \multirow{2}{*}{Class-attribute prompts} & \multicolumn{3}{c}{Prompt template} \\
\cmidrule(l){3-5}
& & \{class\} & A photo of a \{class\}. & A \{attribute\} photo of a \{class\}.\\
\midrule
\multirow{4}{*}{ImageNetV2} & none & 49.39 & 51.34 & \textbf{52.10} \\
& Ours (top 5) & \textbf{53.33} & 52.74 & 52.97\\
& Ours (top 10) & 53.40 & \textbf{53.62} & 53.59 \\
& Ours (top 100) & 53.95 & \textbf{54.06} & 53.78 \\
\midrule
\multirow{4}{*}{ImageNet-Sketch} & none & 31.57 & 33.30 & \textbf{35.31} \\
& Ours (top 5) & 33.12 & 33.65 & \textbf{35.28} \\
& Ours (top 10) & 33.65 & 34.04 & \textbf{35.75} \\
& Ours (top 100) & 34.27 & 34.55 & \textbf{36.04} \\
\midrule
\multirow{4}{*}{ImageNet-A} & none & 20.92 & 21.61 & \textbf{22.58} \\
& Ours (top 5) & 22.78 & 22.86 & \textbf{23.65}\\
& Ours (top 10) & 23.43 & 22.45 & \textbf{23.76} \\
& Ours (top 100) & \textbf{24.00} & 23.16 & 23.78 \\
\midrule
\multirow{4}{*}{ImageNet-R} & none & 58.18 & 56.20 & \textbf{59.29}\\
& Ours (top 5) & 56.29 & 55.46 & \textbf{58.63} \\
& Ours (top 10) & 57.49 & 56.30 & \textbf{59.66}\\
& Ours (top 100) & 59.05 & 57.69 & \textbf{60.44}\\
\bottomrule
\end{tabular}
}
\vspace{1mm}
\caption{\textbf{Domain attributes discovered from unlabeled images make useful prompt templates.}
Top-1 accuracy, CLIP-RN50.
Rows: class-specific attribute prompts (none = plain class name).
Columns: prompt templates; the rightmost ensembles the top-5 rising domain attributes of each dataset (``A cartoon photo of a \{class\}'' on ImageNet-R).
The domain template wins in most settings, by up to 3.7 points over the plain class name on ImageNet-Sketch, with ImageNet-A top-100 and two ImageNetV2 rows as exceptions.
\textbf{Bold} marks the best template per row.}
\label{tab:domaintemplate}
\end{table}

\subsection{Case study: color shift breaks CLIP zero-shot}
\label{sec:fruit}

\begin{figure}[!t]
\begin{center}
   \includegraphics[width=0.9\linewidth]{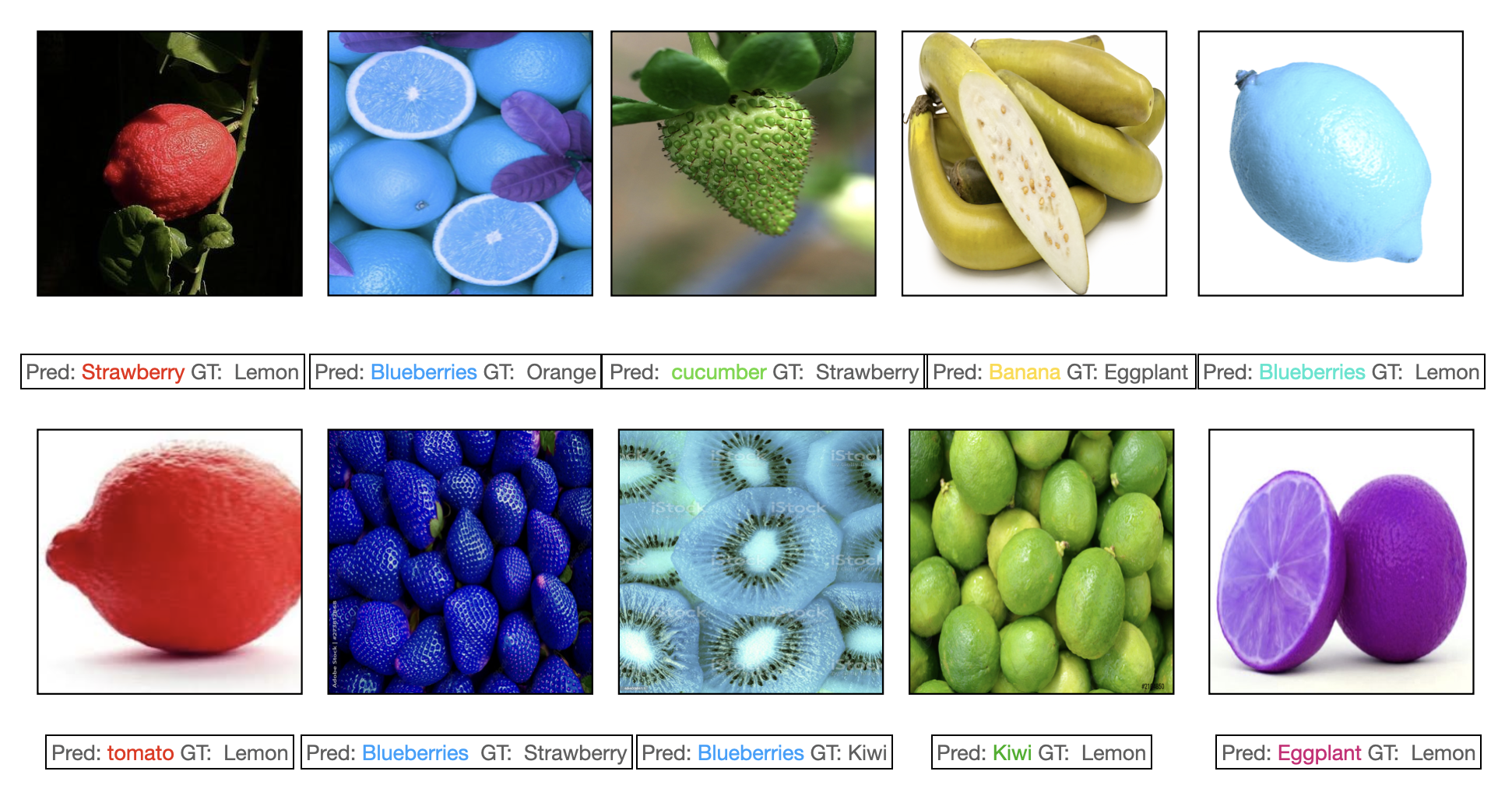}
\end{center}
    \vspace{-2em}
   \caption{\textbf{CLIP zero-shot predictions latch onto color.}
   On color-variant fruit images, zero-shot CLIP often predicts a class that matches the color rather than the object, e.g.\ a yellow eggplant drifts toward lemon.}
\label{fig:clip_color}
\end{figure}

{\bf Motivation.}
CLIP's zero-shot predictions can latch onto color rather than object identity: presented with a yellow eggplant, CLIP drifts toward other yellow concepts such as lemon (Figure~\ref{fig:clip_color}).
To measure this failure mode in isolation we collected a color-variant test set for fruit classification.

\begin{figure}[t]
\begin{center}
   \includegraphics[width=0.8\linewidth]{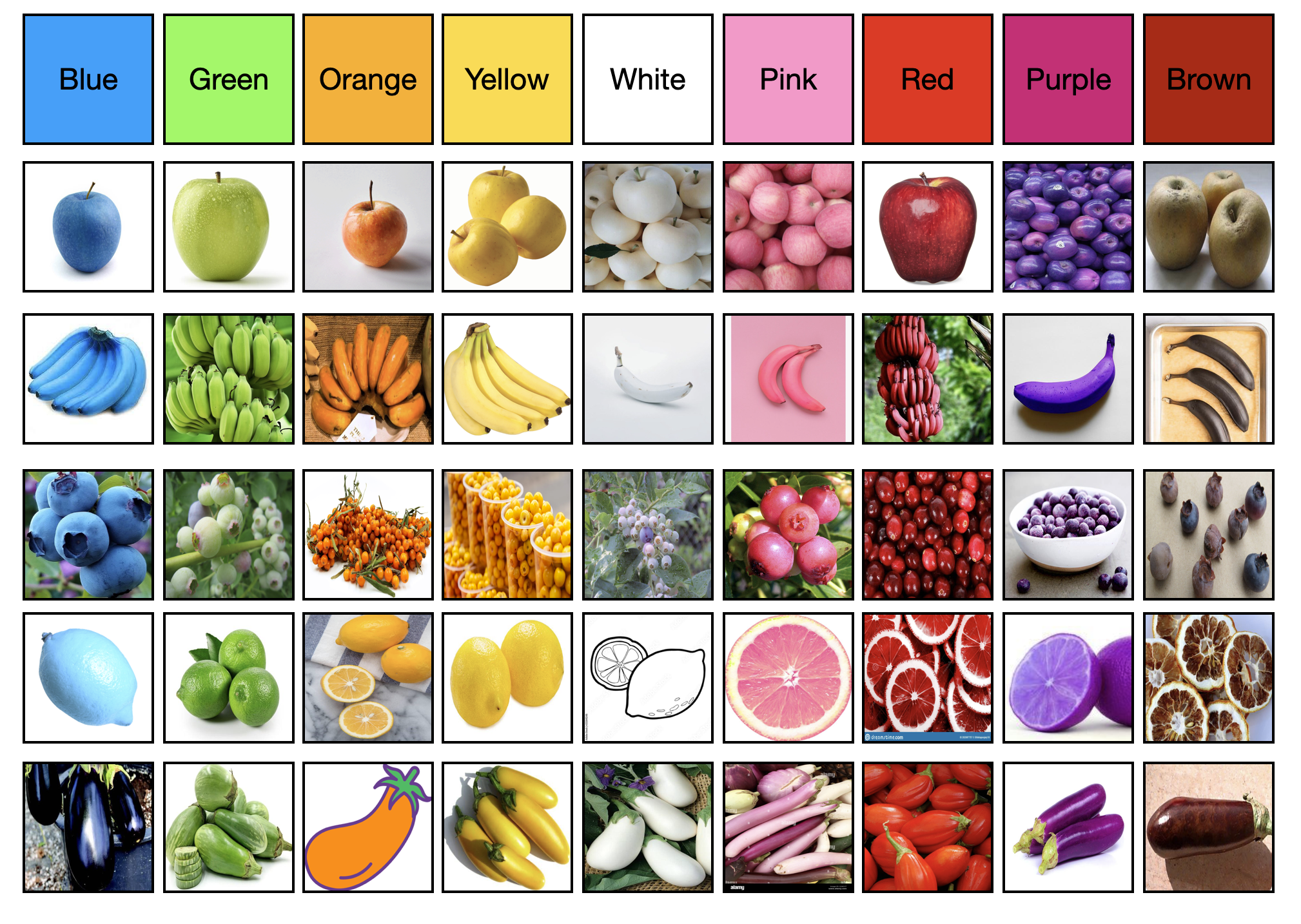}
\end{center}
    \vspace{-2em}
   \caption{\textbf{Examples from the collected color-variant fruit test set.}
   The set spans natural rarities (yellow eggplant), style shifts (sketches and drawings), and artificial colorings (blue and purple lemons), over the 37 classes of the source dataset~\cite{fruits2020kaggle}.}
\label{fig:fruit}
\end{figure}

{\bf Dataset.}
The source is a public fruit dataset~\cite{fruits2020kaggle} of 37 classes and 4{,}440 images (100 train, 10 validation, and 10 test images per class).
We asked ChatGPT to group the 37 class names by color, yielding 9 colors, formed all 333 color-class phrases (``green apple'', ``purple lemon'', ...), crawled 20 web images per phrase, and manually removed irrelevant results.
Merging the color variants by source class gives an additional color-variant test set of the same 37 classes (6{,}050 images overall; examples in Figure~\ref{fig:fruit}).

{\bf Result.}
Zero-shot CLIP drops from 81.02\% on the standard test set to 51.25\% on the color variants, a 30-point fall from color shift alone (Table~\ref{tab:fruit}).
Our attribute prompts improve both settings, to 85.63\% and 55.09\% respectively.
The gain is real but the drop is not repaired; robustly classifying counter-stereotypical colors remains open, consistent with our position that attributes complement rather than replace class-name prompting (Section~\ref{sec:limitations}).

\begin{table}[t]
\centering
\renewcommand{\arraystretch}{1.1}
\resizebox{0.5\linewidth}{!}{
\begin{tabular}{ll|cc}
\toprule
& Evaluation & Test & Color variant \\
\midrule
CLIP & zero-shot & 81.02 & 51.25\\
Ours & zero-shot & \textbf{85.63} & \textbf{55.09}\\
\bottomrule
\end{tabular}
}
\vspace{1mm}
\caption{\textbf{Color shift costs zero-shot CLIP 30 points; attribute prompts recover 4.}
Top-1 accuracy on the fruit dataset's standard test set and our color-variant test set, CLIP-RN50.
\textbf{Bold} marks the better method per column.}
\label{tab:fruit}
\end{table}

\subsection{Vocabulary the text encoder cannot parse}
\vspace{-0.8em}
Some class vocabularies defeat prompt-based classification outright: on iNaturalist~\cite{van2018inaturalist}, whose classes are scientific names such as \emph{Cristidiscoidea}, zero-shot CLIP reaches 3\% accuracy because the text encoder has no grounding for the names.
Selecting attributes from a few images per class replaces the unparseable name with a set of visual words, raising accuracy to 7.25\% with a general-purpose pool of roughly 600 largely human-centric attributes~\cite{Pham2021LearningWild}.
The absolute number stays low and the pool is poorly matched to fine-grained species; the point is the mechanism, which needs no expert to translate jargon into CLIP's vocabulary.

\subsection{Attribute-guided localization}
\begin{figure}[!t]
\begin{center}
 \includegraphics[width=0.83\linewidth]{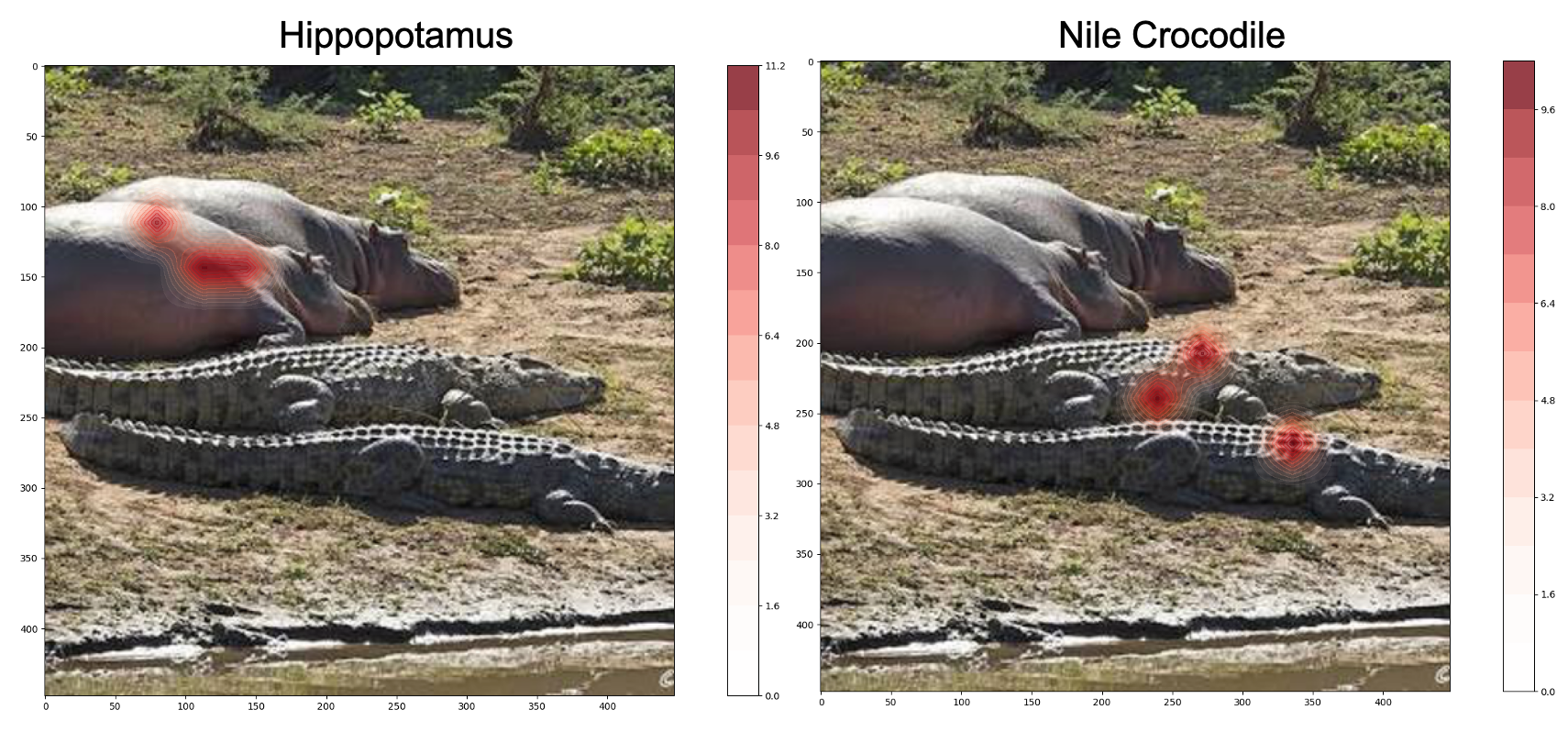}
\end{center}
    \vspace{-2em}
   \caption{\textbf{Class-name-free localization from selected attributes.}
   We remove the attention pooling of CLIP RN50x64, compute cosine similarity between each 14$\times$14 patch embedding and a class's selected attribute texts, and display the resulting heatmap.
   Querying with the hippopotamus attribute set (left) and the Nile crocodile attribute set (right) localizes each animal; the queries contain no class names.
   Warmer color denotes higher mean similarity to the attribute set.}
    \vspace{-1em}
\label{fig:detect}
\end{figure}

\vspace{-0.8em}
Removing CLIP-ResNet's attention pooling preserves the spatial grid, so attribute similarities can be computed per patch.
Querying the patch grid with a class's selected attributes, with no class name in the query, highlights the image regions supporting those attributes and localizes the class (Figure~\ref{fig:detect}).
One forward pass yields the heatmaps for any number of attribute queries, and the same mechanism describes individual regions in words, which we leave as qualitative evidence here.

\vspace{-0.8em}
\section{Limitations}
\label{sec:limitations}
\vspace{-0.7em}

Attribute-only accuracy remains far below class-name prompting (23.80 vs 60.80 on ImageNet with five attributes per class), so when a parseable class name exists, attributes complement it rather than replace it.
In few-shot classification the method plateaus at 61.4\% while CoOp reaches 62.95\% at 16 shots; selection has no capacity to absorb data beyond ranking the pool, and we do not claim the medium-shot regime.
The attribute features are more statistically correlated than the CLIP embedding they re-express (Figure~\ref{fig:feature_disen}), which bounds how independently individual attribute scores can be read.
Our quantitative results use a single backbone (CLIP-RN50) and a selection pool inherited from general-purpose attribute datasets; the iNaturalist experiment (7.25\% absolute) shows what a mismatched pool costs on fine-grained domains.

\vspace{-0.7em}
\section{Conclusion}
\vspace{-0.7em}

Descriptor methods promise interpretable zero-shot classification, but their evidence rides the class name: removed from the prompt, LLM-generated descriptors collapse from 59.5\% to 15.5\% on ImageNet.
Conditioning attribute selection on the target images repairs this at essentially zero training cost, and a matched-pool comparison shows the selection mechanism itself accounts for the repair.
The resulting attribute sets are accurate enough to beat prompt tuning in the extreme few-shot regime and, because they are functions of image collections, they describe datasets, shifts, unparseable vocabularies, and image regions in words.
The finding suggests a protocol change beyond our method: any work that claims attribute-level interpretability from descriptor prompts should report class-name-free accuracy, since the standard protocol cannot distinguish attribute evidence from prompt ensembling.
For concept-bottleneck pipelines, it argues that concept pools should be selected against the deployment distribution rather than generated from labels.

\bibliographystyle{splncs04}
\bibliography{references_fixed,references}

\appendix





\noindent \textbf{\LARGE Supplementary Material}
\\

\titlerunning{Attributes Should Come from Images, Not Class Names (Supplementary)}

\authorrunning{F. Author and S. Author}


\noindent This supplementary material provides the experimental details referenced from the main paper (Section~\ref{sec:supp-setup}), a linear-probing experiment measuring how much of CLIP's representation power the attribute basis preserves (Section~\ref{sec:supp-linearprobe}), and a discussion of editability and attribute-pool relevancy (Section~\ref{sec:supp-discussion}).

\section{Experimental Setup Details}
\label{sec:supp-setup}

{\bf Models.}
All quantitative results use CLIP with the ResNet-50 backbone; the localization figure uses CLIP RN50x64 because its larger 14$\times$14 penultimate grid gives a finer heatmap.
Both encoders stay frozen in every experiment.

{\bf Attribute pool and preprocessing.}
The candidate pool combines the attribute vocabularies of VAW (634 attributes after processing) and LSA, plus the GPT-3 descriptors released by Menon and Vondrick when the protocol calls for their pool.
Preprocessing removes strings that contain explicit class names, exact duplicates, and grammatical punctuation.
Each attribute is encoded bare, with no prompt template and no class name, for the reason given in the main paper: a shared template or class name would correlate all of a class's attribute scores.

{\bf Scoring and pairing.}
Attribute features are scaled cosine similarities with temperature $\tau = 100$.
Class-attribute weights come from a linear probe trained on the attribute features (classification experiments; the probe's per-class weight rows plus bias rank the pool) or from the per-class mean attribute feature (distribution-level applications, which need no labels or training).
Linear probes are trained with AdamW (learning rate $10^{-4}$, weight decay $0.01$) on a single RTX 3090 GPU; fitting completes in under a minute for every setting in the paper.

{\bf Evaluation protocols.}
The class-name protocol follows Menon and Vondrick exactly, including their grammatical modifier: the prompt is ``\{class name\}, which is/has/\{...\} \{attribute\}'' chosen by the attribute's leading word.
The attribute-only protocol reduces the prompt to the bare attribute string.
A test image is scored by the mean similarity to each class's selected attribute prompts, and top-1 accuracy is reported everywhere.

\section{Representation Power of the Attribute Basis}
\label{sec:supp-linearprobe}

The main paper's mutual-information analysis shows the attribute features are semantically structured; this section measures how much of CLIP's discriminative power the change of basis preserves.
We re-express each image as its attribute-similarity vector, keeping the top-1024 attributes to match the dimensionality of the CLIP-RN50 embedding, and linear-probe both representations under identical settings on full ImageNet.

Table~\ref{tab:supp-linearprobe} reports three attribute pools: the VAW vocabulary, the joint (VAW + LSA + LLM) pool, and a control pool of random strings.
The joint pool lands within 2.0 points of the raw CLIP embedding on ImageNet (71.05 vs 73.10), so the interpretable basis costs little linear separability.
The pool ordering (random $<$ VAW $<$ joint) shows that the vocabulary's semantics and diversity both matter for representation quality: random strings retain non-trivial accuracy, as arbitrary projections of a strong embedding do, but trail the joint pool by 5.7 points on ImageNet and by more under shift.

\begin{table}[t]
\centering
\renewcommand{\arraystretch}{1.1}
\resizebox{0.85\linewidth}{!}{
\begin{tabular}{l|ccccc}
\toprule
Representation & ImageNet & -V2 & -Sketch & -A & -R \\
\midrule
Attribute basis, random-string pool & 65.37 & 54.16 & 24.26 & 13.51 & 37.26 \\
Attribute basis, VAW pool & 69.83 & 58.19 & 27.65 & 13.88 & 42.61 \\
Attribute basis, joint VAW+LSA+LLM pool & 71.05 & 59.92 & 30.36 & 16.09 & 46.61 \\
\midrule
CLIP embedding (upper bound) & {\bf 73.10} & {\bf 61.58} & {\bf 33.04} & {\bf 18.69} & {\bf 52.15} \\
\bottomrule
\end{tabular}
}
\vspace{1mm}
\caption{\textbf{The attribute basis preserves most of CLIP's linear separability, and semantic pools beat random strings.}
Top-1 accuracy of a linear probe trained on ImageNet, evaluated on ImageNet and its shifted variants; probes on attribute-similarity features (top-1024 attributes, matching the 1024-d CLIP-RN50 embedding) vs the raw CLIP embedding.
The joint pool sits 2.0 points below the CLIP upper bound on ImageNet; the random-string control trails the joint pool by 5.7 points, so vocabulary semantics matter for the representation.
{\bf Bold} marks the best result per column.}
\label{tab:supp-linearprobe}
\end{table}

\section{Discussion}
\label{sec:supp-discussion}

{\bf Editability and intervention.}
Because a class's representation is a ranked list of named attributes, the classifier admits test-time intervention in the spirit of concept bottleneck models~\cite{koh2020concept}: a practitioner can inspect the word list, delete or reweight an attribute that should not matter, or substitute one that should (replacing ``yellow'' with ``purple'' to describe an unusually colored lemon).
We report this as a property of the representation, not a validated capability: manual swap edits in early experiments gave small gains that did not close the color-shift gap of the main paper's fruit case study, and systematic intervention protocols in the style of CBM interventions remain future work.

{\bf Attribute-pool relevancy.}
The interpretability and accuracy of our method hinge on the relevancy of the attribute pool.
Fine-grained domains need fine-grained vocabulary: the main paper's iNaturalist experiment improves zero-shot accuracy from 3\% to 7.25\% with a largely human-centric pool of roughly 600 attributes, and a species-focused vocabulary should close more of the remaining gap.
The failure mode is mild because selection is cheap: a mismatched pool costs accuracy but never requires retraining, and enlarging the pool toward a universal vocabulary (in the limit, an English dictionary) lets the data mine whatever is relevant.
We leave the universal-pool experiment to future work.



\end{document}